\documentclass[twoside,11pt]{article}
\usepackage{jair, theapa, rawfonts}

\usepackage{graphicx}
\usepackage{subcaption}
\usepackage{amsmath,amssymb,amsfonts}
\usepackage{algorithm}
\usepackage{algpseudocode}

\usepackage{booktabs}
\usepackage{url}

\jairheading{79}{2024}{1313-1341}{08/2023}{04/2024}
\ShortHeadings{Bt-GAN}
{Ramachandranpillai, Sikder, Bergstr\"om \& Heintz}
\firstpageno{1313}

\begin{document}
	
	\title{Bt-GAN: Generating Fair Synthetic Healthdata via Bias-transforming Generative Adversarial Networks}
	
	\author{\name Resmi Ramachandranpillai \email r.ramachandranpillai@northeastern.edu \\\email 
		       \name Md Fahim Sikder \email md.fahim.sikder@liu.se \\
		       \name David Bergstr\"om \email david.bergstrom@liu.se \\
		       \name Fredrik Heintz \email fredrik.heintz@liu.se \\
		       \addr Department of Computer and Information Science (IDA),\\ Link\"oping University, Sweden}
	
	
	\maketitle

	\begin{abstract}
		Synthetic data generation offers a promising solution to enhance the usefulness of Electronic Healthcare Records (EHR) by generating realistic de-identified data. However, the existing literature primarily focuses on the quality of synthetic health data, neglecting the crucial aspect of fairness in downstream predictions. Consequently, models trained on synthetic EHR have faced criticism for producing biased outcomes in target tasks. These biases can arise from either spurious correlations between features or the failure of models to accurately represent sub-groups. To address these concerns, we present Bias-transforming Generative Adversarial Networks (Bt-GAN), a GAN-based synthetic data generator specifically designed for the healthcare domain. In order to tackle spurious correlations (i), we propose an information-constrained Data Generation Process (DGP) that enables the generator to learn a fair deterministic transformation based on a well-defined notion of algorithmic fairness. To overcome the challenge of capturing exact sub-group representations (ii), we incentivize the generator to preserve sub-group densities through score-based weighted sampling. This approach compels the generator to learn from underrepresented regions of the data manifold. To evaluate the effectiveness of our proposed method, we conduct extensive experiments using the Medical Information Mart for Intensive Care (MIMIC-III) database. Our results demonstrate that Bt-GAN achieves state-of-the-art accuracy while significantly improving fairness and minimizing bias amplification. Furthermore, we perform an in-depth explainability analysis to provide additional evidence supporting the validity of our study. In conclusion, our research introduces a novel and professional approach to addressing the limitations of synthetic data generation in the healthcare domain. By incorporating fairness considerations and leveraging advanced techniques such as GANs, we pave the way for more reliable and unbiased predictions in healthcare applications.
	\end{abstract}
	
	\section{Introduction}
	
	Clinical Decision Support Systems \shortcite{rieke2020future} are important for healthcare organizations to improve care delivery in the era of value-based healthcare, digital innovation, and big data. The adoption of advanced artificial intelligence technology in healthcare systems is gathering interest from many researchers \shortcite{ghassemi2021false}, \shortcite{jiang2017artificial}. One example is called precision medicine – predicting which treatment procedures are likely to succeed on a patient based on numerous traits and the treatment context – is the most prevalent application of classical machine learning in healthcare. These systems can yield benefits in terms of improved accuracy, diagnosis, finding new knowledge about the illness conditions and their progression, and the ability to provide patients with a more concrete prognosis and treatment plan by the use of historical data in model development.
	
	Data analysis on Electronic Healthcare Records (EHR) \cite{evans2016electronic} is greatly affected by rules such as the Health Insurance Portability and Accountability Act (HIPPA) \shortcite{edemekong2018health} in the US and the General Data Protection Regulation (GDPR) \cite{regulation2016regulation} in EU, that preserve patient’s privacy. Healthcare organizations are now jointly accountable for any personal data breach since they are responsible for managing all personal data storage and processing in both their organization and that of their suppliers. Over the last three years, Privacy Impact Assessments (PIAs) \cite{wright2012state} have become commonplace in the healthcare industry. GDPR has now placed the delivery of PIAs in the public domain, boosting transparency and information risk ownership clarity.

	Synthetic data generation \shortcite{armanious2020medgan,yale2020generation} tries to preserve the privacy of patients by producing realistic samples to perform downstream tasks. Generative Adversarial Networks have drawn much attention, however, are often criticized for producing low-quality and less diverse samples\shortcite{adiga2018tradeoff}. The state-of-the-art (SOTA) synthetic healthcare data generation methods \cite{armanious2020medgan,edemekong2018health} have explored the concepts of accuracy, utility, and privacy, but paid less attention to fairness in the Data Generation Process (DGP) and in the subsequent tasks. To facilitate the development of equitable analysis and predictions, the DGP should ensure that the synthetic EHR is fair along with other dimensions (such as utility, privacy, etc.). 
	
	There are malignant feature correlations in the medical data, which we call correlation biases. GANs amplify these spurious correlations as studied in \shortcite{gupta2021transitioning}. Another cause of unfairness can be in the direction of fair resemblance - accurately capturing the diversities in sub-groups. We would like to mention \shortcite{bhanot2021problem}, as it raises some concerns about the trustworthiness of synthetic data in the direction of fair resemblance. We use the term \textit{representation fairness} instead of \textit{fair resemblance} throughout this study as this refers to how different sub-group proportions are represented in the synthetic data as compared to real data. Therefore representation biases are seen when the sub-groups are missing, underrepresented, or overrepresented in the synthetic data.

\begin{figure}[]
			\centering
			\includegraphics[width=0.6\linewidth, height=4cm]{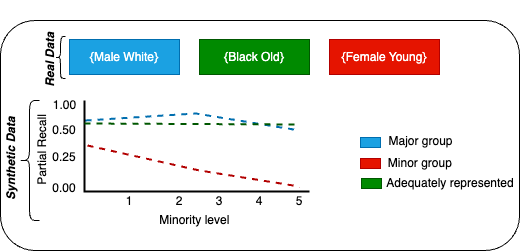}
			\caption{An example of partial recall for groups with underrepresentation (minor), overrepresentation (major), and adequate representation in data generation.}
			\label{fig:partial-recall}
		\end{figure}

Figure \ref{fig:partial-recall} illustrates a simplified example wherein the degree of representation in the synthetic data is measured using Partial Recall \shortcite{kynkaanniemi2019improved}. The generated distribution shows a substantial recall discrepancy between majority and minority groups, notably expanding as the minority level increases. Consequently, the synthetic data with minority samples exhibit both inadequate quality and coverage problems, potentially resulting in incidental or spurious correlations within the synthetic dataset.

	In this paper, we address the problem of synthetic healthcare data fairness; generate fair data from biased data (correlation biases), and promote representation fairness in the target data.
	
	\textbf{Contributions}. The problem definition, design, analysis, and experimental assessment of the proposed framework are the main contributions of this research. Particular contributions include:
	
	\begin{itemize}
		\item Development of a principled GAN framework for synthetic healthcare data generation with guaranteed fairness in the target in contrast to existing techniques such as HealthGAN and MedGAN, where no fairness is ensured. 
		\item Problem definition considering how the DGP is affected by various biases from the data (correlation bias) and how the generative model injects additional biases (representation biases) in the training process. 
		\item A comprehensive experimental assessment of the proposed framework on health database and comparison with state-of-the-art methods in terms of data utility and fairness.
		\item An analysis of the bias amplification \shortcite{wang2019balanced} of the proposed approach and comparison with the state-of-the-art.
		\item An explainability analysis using SHAP \shortcite{nohara2019explanation} to validate the trustworthiness of the proposed framework. 
	\end{itemize}
	
	\section{Related Works}
	\noindent We focus on the related literature in terms of 
	i) synthetic data generation in healthcare systems, ii) fair data generation, and (iii) representation issues in GANs in contrast to fairness measures, which we define in Section \ref{sec:desiderata}.  
	
	{\bf GANs in Healthcare Systems.} Synthea \shortcite{walonoski2018synthea} simulates patient records from birth to the present day using modules informed by clinicians and real-world statistics. It claims to preserve utility by employing healthcare practitioners and real statistics to construct rules that synthesize the data, which assures privacy. HealthGAN \cite{yale2020generation} generates synthetic health data of multivariate nature. The loss function is based on the Wasserstein distance \shortcite{arjovsky2017wasserstein} and they used \textit{Synthetic Data Vault (SVD)} \shortcite{patki2016synthetic} for categorical encoding. Medical Image translation, MedGAN \cite{armanious2020medgan}, is an end-to-end framework by merging the adversarial network with non-adversarial losses. The discriminator network is a pre-trained feature extractor, which penalizes the discrepancy. Among the models mentioned above, HealthGAN provides promising results in terms of accuracy and utility \cite{bhanot2021problem}.
	
	{\bf GANs in Fair Data Generation.} The Fairness aware GAN (Fair GAN) \shortcite{xu2018fairgan} approach trains a generator to produce fair representations by an additional discriminator that knows the real distribution of the protected features in the training data. DECAF \shortcite{van2021decaf} is designed to generate fair synthetic tabular data with an assumption that the underlying causal structure is known.  It uses individual generators to generate features sequentially based on the causal graph, while de-biasing is done at the inference time. 
	
	{\bf Improving Under-representation in GANs.} GANs frequently experience mode collapse and generate samples with low diversity because of the unstable nature of the min-max game between a generator and a discriminator. Better data coverage has been advocated using objective function-based methods \shortcite{kodali2017convergence}, \shortcite{mao2017least} and structure-based methods \shortcite{radford2015unsupervised}, \shortcite{zhang2019self}. Even while they work well to increase data coverage overall, these methods don't give minor modes any extra attention. They frequently fail to recover minor modes when the minority ratio for a given feature is meager. Techniques such as label smoothing have been successfully applied to increase the performance of GANs when the scale of the training data is limited \shortcite{Zheng_2017_ICCV}. Another category of methods is sampling-based methods \shortcite{lee2021self} which give extra attention to minor modes and then promote these modes by score-based sampling. Discriminator Rejection Sampling (DRS) \shortcite{azadi2018discriminator} is proposed to apply rejection sampling to filter the synthetic samples based on density ratio estimation. Likewise, GOLD \shortcite{mo2019mining} reweighted fake samples to improve the representation (or resemblance). Top-k training \shortcite{sinha2020top} modifies the generator using only the top-k synthetic samples to promote representation. Similar to DRS, Dia-GAN \cite{lee2021self} proposed a discrepancy score based on the empirical mean and variance over multiple epochs. We follow a similar approach to tackle representation biases in data generation.
	
	In addition, none of the above methods have successfully handled partially unlabelled data. Medical data often contain missing labels which if not treated properly can adversely affect a patient's health.

	In summary, we present Table \ref{tab:comparison-related}  to compare the methods with different key areas of interest. As far as we know, our Bt-GAN is the first GAN architecture that tackles correlation biases, representation biases, and partially unlabeled data in an end-to-end framework in the underlying DGP.
	\begin{table*}[!h]
		\centering
		{\small
				\begin{tabular}{lccccc}\toprule
					\textbf{GAN}     &\textbf{a}    &\textbf{b}    &\textbf{c}    & \textbf{Method}   &\textbf{Goal} \\ 
					\midrule
					HealthGAN  &No    &No     &No   & Wasserstein-     &Synthetic    \\
					\cite{yale2020generation}&&&&distance&health data  \\
					MedGAN   &No    &No     &No   & Progressive-    &Synthetic     \\
					\cite{armanious2020medgan}&&&&refinement &health data  \\
					FairGAN  &Yes   &No    &No     &Adversarial-     &Fair synthetic  \\
					\cite{xu2018fairgan}&&&&de-biasing  &data\\
					DECAF   &Yes    &No    &No     &Causal structure    & Fair synthetic \\
					\cite{van2021decaf}&&&&&tabular data\\ 
					\midrule
					Bt-GAN(ours) &Yes   &Yes   &Yes    &bias-transforming DGP + & Fair and representative\\
					& & & & score-based sampling  &  synthetic healthdata\\
					\bottomrule
				\end{tabular}
			}
			\caption{Comparison of related works with different key areas of interests: (a) Correlation bias, (b) Representation bias, and (c) Partially unlabelled data.}
			\label{tab:comparison-related}
		\end{table*}
		
		\section {Unfairness in Data Generation Process}
		According to \shortcite{wachter2020bias}, the term \textit{bias-preserving} in fairness literature means that the status quo (or training dataset) is a baseline and the model tries to reproduce the historic performances in the status quo, which only accounts for direct discrimination. But, the \textit{bias-transforming} metrics address indirect discrimination by fixing the structural inequalities in the status quo. In the following, we describe the causes of unfairness in DGP in detail and illustrate how we approach the problem definition with the concepts studied in \cite{wachter2020bias}.
		\subsection{Learning of Protected Attribute Information}
		Learning protected attribute information is one of the key elements of unfairness in classification or prediction, as described in the literature \shortcite{creager2019flexibly}, \shortcite{locatello2019fairness}, \shortcite{song2019learning}. This situation can be worse in synthetic data generation using GANs as it amplifies the existing biases in the DGP as studied in \cite{gupta2021transitioning}.
		
		Suppose we have an ideal biased dataset $\mathcal D_{bias}=\{\mathcal {\Tilde{X}}, {\mathcal {\Tilde{Y}}}, {\mathcal{\Tilde{S}}}\}$ where each label $\mathcal {\Tilde{Y}}$ is correlated with each sensitive attribute in equal intensity.  Then for a well-trained fixed discriminator\footnote{we assume that the data used to train is very large and the model has enough capacity}, the generative model, once converged, captures the dependencies from $\mathcal D_{bias}$, which means the labels in the synthetic data are also correlated with the sensitive attribute in equal intensity. Based on this and motivated by \cite{wachter2020bias}, we define the following in the context of GANs and correlation bias: Let $\mathcal D$ be the real dataset containing correlation biases and $\hat{\mathcal D}$ be the synthetic data generated by the underlying generator $G$ of a GAN.
		
		\textbf{Definition 1 - Bias-preserving DGP (Bp-DGP)}. \textit{A DGP is bias-preserving if and only if the underlying generative model $G$, once optimized, learns to obtain a transformation from a Multivariate Normal Distribution (MVD) to the real data distribution by preserving the exact correlations from $\mathcal D$ and replicate it in $\hat {\mathcal D}$} across sensitive groups.  
		

		\textit{Remark} - A Bp-DGP seeks to keep the historic performances from the real data in the output of a target model when trained with synthetic data with equivalent error rates for each group as shown in the real data.
		
		
		\subsection{Representation Bias: Failure of GANs in Capturing the Exact Representations }
		The synthetic generation based on GANs fails to capture the exact sub-group proportions from the real data as GANs try to match the distributions of real data at the dataset level. We define the following in the context of GANs and representation biases:
		
		\textbf{Definition 2 - Density-preserving DGP (Dp-DGP)}. \textit{A DGP is said to be density preserving if and only if the underlying generative model $G$, once optimized is learned to generate synthetic data $\hat{\mathcal D}$ in such a way that the ratio of sub-groups between $\mathcal D$ and $\hat{\mathcal D}$ should be the same.}

\textit{Remark} - Due to the min-max objective of GAN optimization, the DGP can inject additional biases in the form of representation biases due to the mode collapse problem as studied in \cite{gupta2021transitioning}.

Correlation biases and representation biases are critical in synthetic data generation based on GANS.

  		\section{Desiderata}
		\label{sec:desiderata}
		\subsection{Fairness Definition}
		Formally, let $\mathcal D=\{\mathcal X, \mathcal S, \mathcal Y\}$ be a dataset containing biases, where $ X \in \mathcal X \subset \mathbb{R}^d$ is a random variable of non-sensitive features, $S\in \mathcal S $ be sensitive features and $Y \in \mathcal Y $, the labels. Also, let $\mathcal U(\mathcal S, \mathcal Y)$ be a definition of algorithmic fairness. We define the following algorithmic fairness measure to eliminate indirect discrimination \cite{wachter2020bias}:
		
		{\bf Definition 3 - (Statistical Parity }\shortcite{dwork2012fairness}). \textit{ Suppose we have a function $h: X \rightarrow Y^{\prime}, Y^{\prime}=\{0,1\}$ for binary classification, and assume $S$ splits $X$ into a majority set $\mathcal{M}$ and a minority set $\mathcal{M}^{\prime}\left(X=\mathcal{M} \cup \mathcal{M}{ }^{\prime}\right)$, then the function $h$ satisfies statistical parity if $P[h(x)=1 \mid x \in \mathcal{M}]=P\left[h(x)=1 \mid x \in \mathcal{M}^{\prime}\right]$, where $x$ denotes an instance of $X$ and $P[.]$ denotes the probability of an instance.}
		
		We assume the protected attribute is binary for notational convenience as this can be extended to non-binary classification.
		
		\subsection{Mutual Information}
		\label{sub:mi}
		
		{\bf Definition 4 - {(Mutual Information}} \shortcite{veyrat2009mutual}){\bf .} \textit{Let $W_{1}$ and $W_{2}$ be two random variables, the product of marginal distribution be $p_{W_{1}}$ $p_{W_{2}}$ and the joint distribution be $p_{W_{1}, W_{2}}$, then the mutual information between $W_{1}$ and $W_{2}$ can be defined as:} 
		
		\begin{equation}
			\begin{split}
				\centering
				\label{eq:eq1}
				I(W_{1};W_{2}) & = H(W_{1}) - H(W_{1}\mid W_{2}) =\\
				& \qquad \int_{W_{1}} \int_{W_{2}} p_{(W_{1},W_{2})} \log \frac{p_{(W_{1},W_{2})}}{p_{(W_{1})} p_{(W_{2})}} \,dW_{1}.\ d W_{2}
			\end{split}
		\end{equation}
		Unlike correlation coefficients \shortcite{benesty2009pearson} (which could only estimate the linear dependence), mutual information is used to measure linear as well as non-linear dependencies between two random variables.

		{\bf Definition 5 - {(Relationship between Statistical Parity and Zero Mutual Information }}\shortcite{ghassami2018fairness}) {\bf .} \textit{Given a predicted outcome, $Y^{\prime}=\{0,1\}$, and a protected attribute $S$, then,}
		
		\begin{equation}
			p\left[Y^{\prime} \mid S\right]=p\left[Y^{\prime}\right] \Leftrightarrow p_{Y^{\prime}, S}=p_{Y^{\prime} } p_{S} \Leftrightarrow I\left(Y^{\prime}; S\right)=0
		\end{equation}
		If the two random variables are independent of each other we get zero mutual information.  
		\subsection{Generative Adversarial Networks(GAN)}
		The Generative Adversarial Network (GAN) is a prominent member of the generative models family, comprising two essential components: a generator and a discriminator. The generator takes in random noise $z \sim \mathcal{N}(0,1)$ as input and endeavors to produce realistic data, represented as $x \sim P_\mathcal{D}$. On the other hand, the discriminator is responsible for distinguishing between the generated data and the original data. As time progresses, the generator becomes more adept at deceiving the discriminator, while the discriminator strives to differentiate between real and fake data. This dynamic creates a zero-sum game, where both networks engage in a continuous battle until they reach a state of equilibrium known as the Nash equilibrium. To quantify the performance of the generator $(\mathrm{G})$ and discriminator $(\mathrm{D})$, the following loss function is employed:

  \begin{equation}
\begin{split}
    \min_{\mathrm{G}} \max_{\mathrm{D}} & V(\mathrm{G},\mathrm{D}) = E_{x\sim P_{\mathcal{D}}}[log(\mathrm{D}(x))] +\\
    & E_{z\sim p_{z}(z)}[log(1-\mathrm{D}(\mathrm{G}(z)))]
\end{split} 
\end{equation}

		\section{Synthetic Data Fairness}
		Synthetic data fairness means generating fair data from biased data such that any downstream model trained on fair synthetic data will have fair predictions in real data assuming that the underlying prediction model does not possess any explicit biases. 
		
		To achieve synthetic data fairness considering correlation biases (Definition 1) and representation biases (Definition 2) in the DGP, we propose to define Bias-transforming DGP (Bt-DGP) as: 
		
		\textbf{Definition 6 - Bias-transforming DGP (Bt-DGP)}. \textit{Let $G$ be a generative model and $\mathcal U(\mathcal S, \mathcal Y)$ be a definition of algorithmic fairness (such as statistical parity in our case). A DGP is bias-transforming if and only if it is Dp-DGP and the underlying generative model $G$ transforms the existing biases in $\mathcal D$ in such a way that $\hat{\mathcal D}$ is fair, as defined by $\mathcal U(\mathcal S, \mathcal Y)$ and the utility is maintained with respect to any downstream tasks.} 

\begin{figure}[h]
			\centering
			\includegraphics[width=0.6\linewidth, height=11cm]{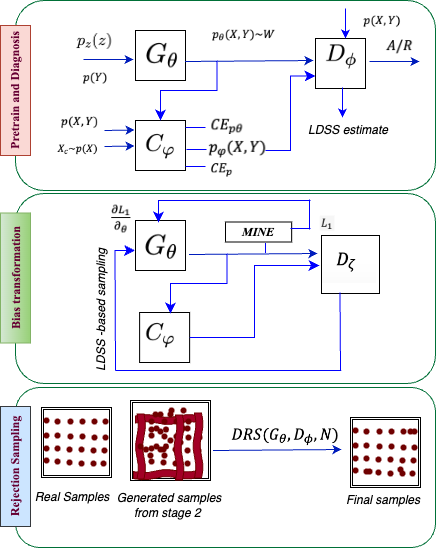}
			\caption{Architecture of Bt-GAN: the utilities of  $\mathrm{C}_{\varphi}, \mathrm{G}_{\theta}$, $\mathrm D_{\zeta}$, and $\mathrm{D}_{\phi}$ are shown. The symbols $A$ and $R$ denote \textit{accept} and \textit{reject} respectively. The discriminator,  $D_{\phi}$ accepts if it thinks it is from the true data distribution denoted as $p$.}
			\label{fig:fig1}
		\end{figure}
  
		\subsection{Problem Definition}
		The Synthetic Data Fairness Problem (SDFP) is to generate fair data $\hat {\mathcal D}=\{ \hat {\mathcal{X}}, \hat {\mathcal{S}}, \hat {\mathcal{Y}\}}$ from $\mathcal D$ through Bt-DGP.

    \textbf{Objective}: To design a framework that accounts for both representation issues and spurious correlations and to guarantee the fairness of any downstream models trained on the synthetic data.
		
\textbf{Approach}: For tackling correlation biases we use and exploit Mutual Information (MI) \shortcite{hemamou2022delivering}, \shortcite{kang2021multifair}, \shortcite{ghassami2018fairness} as described in Section \ref{sub:mi}. For representation biases, we adopt a score-based sampling as detailed in Section \ref{sec:btgan}.

		In the next section, we describe how we achieve Bt-DGP by proposing the Bt-GAN framework.

		
		\section{Bias-transforming Generative Adversarial Networks}
		\label{sec:btgan}
		This section presents our Bias-transforming Generative Adversarial Network (Bt-GAN) framework in detail.

		The whole process of Bt-GAN can be divided into 3 stages: 
		\begin{enumerate}
			\item \textbf{Pre-train and Diagnose} - the generator $G$ of a GAN learns to generate high-quality samples from a large real-world dataset (biased and partially labeled). During this process, we diagnose representation biases by recording the corresponding sub-group densities from synthetic data.
			\item \textbf{Bias-transform} - the pre-trained generator $G$ from stage 1 is fine-tuned to generate fair distribution. This involves unlearning the sensitive correlations using a fairness penalty and enforcing representation fairness using score-based weighted sampling based on densities, computed from stage 1. This transformation stage tackles both correlation bias from the data and the representation bias from the GAN, thereby encouraging GAN to learn from the under-represented regions of the data manifold.  
			\item \textbf{Discriminator Rejection Sampling (DRS)} - The score-weighted sampling in step 2 injects new biases towards under-represented regions. We correct it using rejection sampling \cite{azadi2018discriminator} using the discriminator trained in stage 1. 
		\end{enumerate}
		An overview of Bt-GAN architecture is given in Figure \ref{fig:fig1} and we describe all the stages in detail as follows:
		
		\subsection{Pretrain and Diagnosis}
		Healthcare data often have missing values, especially partially unlabeled. So, to completely capture the true data distribution from the partially unlabeled data, we employ Triple GAN \shortcite{li2017triple}, a semi-supervised GAN framework involving 3 components: i) a classifier $C_{\varphi}$ for the conditional distribution, $p_{\varphi}(y \mid x)$, ii) a Generator $G_{\theta}$ that characterizes $p_{\theta}(x \mid y) \approx p(x \mid y)$, and iii) a Discriminator $D_{\phi}$ that predicts whether the data pair $(x, y)$ is from $p_{\theta}(x \mid y)$ (fake) or from $p(x \mid y)$ (real). After a sample $x$ is drawn, $C_{\varphi}$ produces $p_{\varphi}(x, y)=p(x) p_{\varphi}(y \mid x)$. Then, the joint distribution produced by $G_{\theta}$ becomes, $p_{\theta}(x, y)=p(y) p_{\theta}(x \mid y)$. Let $x$ be obtained by the latent variable $z$, then, $x=$ $G_{\theta}(y, z), z \sim p_{z}(z)$, where $p_{z}(z)$ can be uniform or normal distribution.
		
		The mini-max game for Triple GAN can be formulated as \cite{li2017triple}:
		
		
		\begin{equation}
			\begin{split}
				L_{G C D} & =\min _{\varphi, \theta} \max _{\phi} V\left(C_{\varphi}, G_{\theta}, D_{\phi}\right)=\\
				& \qquad E_{(x, y) \sim p(x, y)}[\log D(x, y)]+\\
				& \lambda E_{(x, y) \sim p_{\varphi}(x, y)}[\log (1-D(x, y))]+\\
				& (1-\lambda) E_{(x, y) \sim p_{\theta}(x, y)}[\log (1-D(G(y, z), y))]+L_{C E},
			\end{split}
		\end{equation}
		
		where, $\lambda \in(0,1)$ is the balance factor that controls the significance of classification and generation, we set $\lambda=0.5$ for the entire training process.
		To make $p(x, y)=p_{\varphi}(x, y)=p_{\theta}(x, y)$, a cross entropy loss $L_{C E}=E_{(x, y) \sim p(x, y)}\left[-\log p_{\varphi}(y \mid x)\right]$ has been added \cite{li2017triple}. 
		
		During the training of the triple GAN, we record the densities for each selected sub-groups. The densities are estimated by a measure called log disparity of sub-groups (LDS) based on density ratios as follows:

		{\bf Definition 7 - (Log Disparity of Sub-groups (LDS))}.\textit{Let $f(x)$ be a membership function for the binary definition of sub-groups $g_{j} \in \mathbb{G}, 1 \leq j \leq|\mathbb{G}|$, then $f(x)=1$, if $\mathrm{x} \in g_{j}$ and $f(x)=0$, if $\mathrm{x} \notin g_{j}$. The log disparity of sub-groups, LDS(.) between $p_{\mathcal{D}}$ and $p_{\hat {\mathcal{D}}}$, is defined as:}
		
		\begin{equation}
			\centering  
			\operatorname{LDS}(x_{i})=\log \left(\frac{o\left(f(x_{i})=1 \mid x_{i} \in g_{j} \in p_{\hat {\mathcal{D}}}\right)}{o\left(f(x_{i})=1 \mid x_{i} \in g_{j} \in p_{\mathcal{D}}\right)}\right) ,
		\end{equation}
		
		where $o(f(x_i) = 1 \mid x_i \in g_j \in p_{\hat {\mathcal{D}}})$  = $P(f(x_i) = 1 \mid  x_i \in p_{\hat{\mathcal {D}}}) / (1 - (P(f(x_i) = 1 \mid  x_i \in g_j \in p_{\hat {\mathcal{D}}})))$, and can similarly be calculated for $o\left(f(x_{i})=1 \mid x_{i} \in g_{j} \in p_{\mathcal D}\right)$. Log disparity of sub-group can be computed for all the available sub-groups by changing $f(x)$ to different $g_{j}$ in $\mathbb{G}$.
		
		The discriminator $\mathrm D_{\phi}$ output can be used to estimate the disparity of sub-groups, $LDS(.)$ between $\mathcal D$ and $\hat {\mathcal D}$.

		\subsection{Bias Transformation}
		In the bias transformation step, the pre-trained GAN learns the fair distribution underlying the chosen attributes. The fair distribution means that the generated data should not contain both correlation biases and representation biases. 
		
		\subsubsection{Tackling Correlation Biases: Unlearning Sensitive Correlations using Information-constrained DGP} 
		GANs learn spurious correlations from data to converge to the true data distribution. Let $W$ be the generated space. Specifically, we define learning of protected attribute information as increasing $I(W; S)$ which is the mutual information between $W$ and $S$ in the DGP.
		We  exploit Definition 5 and formulate a mutual information reduction problem between the generated space $W$, which is a pair $\left\{X^{\prime}, Y^{\prime}\right\}$, and the vector encoded sensitive features, $S$ ($W$ and $S$ are random variables). So, we propose the following information-constrained minimax objective function:
		
		\begin{equation}
			\begin{split}
				\min _{\varphi, \theta} & \max _{\phi}  V\left(C_{\varphi}, G_{\theta}, D_{\phi}\right)= E_{(x, y) \sim p(x, y)}[\log D(x, y)]+\\
				& \lambda E_{(x, y) \sim p_{\varphi}(x, y)}[\log (1-D(x, y))]+\\
				&(1-\lambda) E_{(x, y) \sim p_{\theta}(x, y)}[\log (1-D(G(y, z), y))]+\\
				& L_{C E} ;   \exists I(W ; S)=0,
			\end{split}
		\end{equation}
		The condition, $I(W; S)$ depends on the trainable parameter $\theta$. In this setting, $S$ is a constant vector-encoded representation of sensitive features selected by the data manager (or data owner). 
		
		For estimating the fairness penalty term $I(W; S)=0$, we use Mutual Information Neural Estimation (MINE) \shortcite{belghazi2018mine}, and the maximization can be handled by back-propagation. The loss function $L_{I}$ for MINE is:
		\begin{equation}
			L_{I}=E_{\hat{p}(W, S)}\left[T_{\eta}(W \mid S)\right]-\log E_{\hat{p}_{w}, \hat{p}_{s}}\left[e^{T_{\eta}(W \mid \hat{S})}\right],
		\end{equation}
		
		where $T_{\eta}$ is a statistical neural network. The loss term, $L_{I}$ can provide an estimate of MI, once the parameter $\eta$ is maximized and it depends on $\theta$. Also, $\hat{p}$ denotes the empirical estimation of distribution. Therefore:
		
		\begin{equation}
			L_{M I}=I[\widehat{W ; S}]=\max _{\eta} L_{I}(\eta, \theta)
		\end{equation}
		
		The fairness penalty term helps the generator in learning a distribution $W$ from $Z$ with the constraint of not containing any malignant information related to $S$. 
		
		So, the final loss, $L_{F}$ for Bt-GAN would be:
		\begin{equation}
			L_{F}=\underbrace{\underbrace{L_{G C D}}_{Semi-supervised-generation}+\underbrace{\alpha L_{M I}}_{MI de-biasing}}_{Fair-generation}
		\end{equation}
		
\begin{figure*}[]
			\centering
			\includegraphics[width=0.8\textwidth, height=4cm]{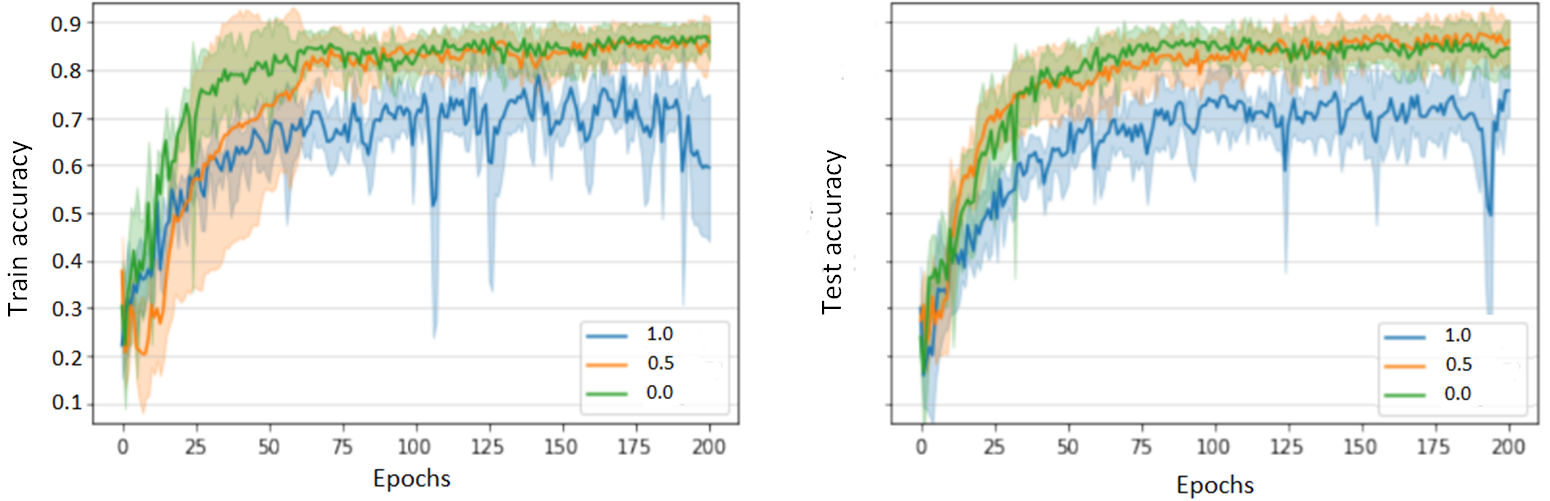} 
			\caption{Ablation study showing the effect of different values of $\alpha$ on test and train accuracy. Note that, when $\alpha=1$, the effect of MI reduction between $W$ and $S$ is large, but the accuracy also drops severely. When $\alpha=0.5$, the model is performing comparatively better on the mortality prediction task. Also, when $\alpha=0.0$, the MI reduction part in equation 8 is inactive and thus is the fairness constraint. According to this, we set $\alpha=0.5$ to balance the quality-fairness trade-off for the entire process.}
			\label{fig:fig2}
		\end{figure*}

		The parameter $\alpha$ balances the MI reduction and the quality of generation. We give an ablation study (Figure \ref{fig:fig2}) on how it affects the performance of the model by varying $\alpha$. The Triple GAN minimax objective is optimized by iteratively modifying the generator, discriminator, and classifier with respective losses. 
		
		\subsubsection{Improving Data Coverage using Density-preserving Sampling}
		Following \cite{lee2021self}, we propose a Log Disparity Sub-group Score (LDSS) (based on LDS in equation 4), which measures how close the real and synthetic distributions on sub-groups for training sample $x$ over $T$ iterations, as:
		\begin{equation}
			LDSS(x_i,T)= \frac{1}{T} \Sigma_{k \in T} LDS(x)_k
		\end{equation}
		
		The aim is to design a sampling probability for an instance $i$, based on LDSS, and propagate it through SGD to design the batch size, $\mathcal{D}_{\mathrm{B}}=x^{k}: x^{k}=x_{i}, i \sim P_{L D S S}(i)$ for $k=$ $1,2, \ldots, B$. ie, each $x_{i} \in \mathcal{D}$ is sampled with a probability of $P_{L D S S}(i)$.
		
		{\bf Definition 8 - ($LDSS$-based Sampling Probability)}. \textit{For a training dataset $\mathcal{D}$, we denote each instance as $x_{i}$  for notational convenience, The sampling probability $P_{L D S S}(i)$ of the $i^{\text {th }}$ data instance is calculated over a number of steps $T$ by:}
		
		\begin{equation}
			\ P_{L D S S}(i)=\frac{\operatorname{LDSS}\left(x_{i}, T\right)}{\sum_{k=1}^{|\mathcal{D}|} \operatorname{LDSS}\left(x_{k}, T\right)}
		\end{equation}
		
		Based on the LDSS, we define the following:
		
		{\bf Definition 9 - ($LDSS$-based Density-preserving DGP)}. \textit{A DGP is LDSS-based Dp-DGP, if and only if the LDSS of sub-groups (for each $g_{j} \in \mathbb{G}$ ) between ${p_{\mathcal{D}}}$ (real data distribution) and $p_{\hat {\mathcal{D}}}$ (synthetic data distribution) is below an acceptable threshold $\delta$}.
		
		
		We elaborate in Section \ref{sub:reprefair} on how the threshold $\delta$ has been set to various levels for evaluating the representation biases in synthetic health data.

		With reference to Definition 9, we re-iterate Definition 6 as:
		
		{\bf Definition 10 - Bias-transforming DGP (Bt-DGP)}. \textit{Let $G$ be a generative model and $\mathcal U(\mathcal S, \mathcal Y)$ be a definition of algorithmic fairness (such as statistical parity in our case). A DGP is said to be bias-transforming if and only if it is LDSS-based Dp-DGP and the underlying generative model $G$ transforms the existing biases in $\mathcal D$ in such a way that the $\hat{\mathcal D}$ is fair, evaluated by a definition of algorithmic fairness, $\mathcal U(\mathcal S, \mathcal Y)$.} 
		
		\subsection{Discriminator Rejection Sampling}
		LDSS-based sampling creates biases as the generated distribution $p_{\hat{\mathcal D}}$ differs from the real data distribution $p_{\mathcal{D}}$. We employ rejection sampling \cite{azadi2018discriminator} and use the discriminator $D_{\phi}$ from stage 1 (as it knows the real distribution) with an acceptance level $p_{\mathcal{D}}(x) / \operatorname{L p}_{\hat{\mathcal D}}(x)$, for some constant $L>0$. We use the same architecture for both discriminators (except for the sigmoid activation). Also, to speed up the processing, the parameters of $D_{\zeta}$ are instantiated with $D_{\phi}$.

		
		

  \section{Theorems and Proofs}
  
\textbf{Theorem 1: } \textit{An ideal data generation process\footnote{a process in which the underlying generative model exactly converges to the true data distribution} is Bp-DGP}.
		
\textbf{Proof}. For a well-trained fixed discriminator of a Triple GAN, $D_\phi$, the global convergence of the optimization equation $L_{GCD}$ is achieved when $p_g(x,y)=p(x,y)=p_c(x,y) $, where $p_g(x,y), p(x,y),$ and $p_c(x,y)$ respectively denote generator distribution, real data distribution, and classifier distribution. Since the tree-player game is a hard constraint compared to two-player optimization, it absolutely captures the real dependence from data $\mathcal D$ even with missing labels. Note that we have not changed the discriminator. Therefore the synthetic data $\hat {\mathcal D}$ contains all the biases in $\mathcal D$, and hence it is Bp-DGP.

\textbf{Theorem 2: }\textit{An ideal data generation process is Dp-DGP}.
		
\textbf{Proof}. Let $G$ be the generator, $D$ be the discriminator, and $C$ be the classifier of a Triple GAN. We assume enough capacity for $G$, $D$, and $C$. Also, we assume that the Generator $G$ does not possess any mode collapse under the ideal DGP and converges when $p_g(x,y)=p(x,y)=p_c(x,y) $ by replicating all the sub-group densities from $\mathcal D$. Therefore, an ideal DGP is Dp-DGP.
		
\textbf{Theorem 3:} \textit{An ideal DGP is both Bp-DGP and Dp-DGP}.
		
\textbf{Proof}. It can be proved by Theorem 1 and Theorem 2.

{\bf Theorem 4:} \textit {Given optimal discriminator $D_{C, G}^{*}(x, y)$, the global minimum of the generator and classifier loss is achieved if and only if $p(x, y)=p_{\phi}(x, y)=p_{\theta}(x, y)$.} 
		
We use the following Theorem to prove Theorem 4.
		
{\bf Theorem 4.1}. \textit{ For any fixed $C$ and $G$, the optimal $D^{*}$ of the game defined by the utility function $L_{G C D}$ is:}
		\begin{equation}
			D_{C, G}^{*}(x, y)=\frac{p(x, y)}{p(x, y)+p_{\gamma}(x, y)}
		\end{equation}
		where: $p_{\gamma}(x, y)=(1-\gamma) p_{g}(x, y)+\gamma p_{c}(x, y)$ is a mixture distribution for $\gamma \in(0,1)$. We refer to Lemma 3.1, Lemma 3.2, and Theorem 3.3  of \cite{li2017triple} for the proof for Theorem 4.
		
\textbf{Proof}. Note that we have made zero changes to the Triple GAN discriminator and therefore we can prove that both $\mathrm{C}$ and $\mathrm{G}$ will converge to the true data distribution\footnote{we assume all the components have enough capacity} if in each iteration the discriminator has been trained to achieve the optimum for a fixed $C$ and $G$ and $C$ and $G$ subsequently modified to maximize the discriminator loss. Since the discriminator is trained optimum, it penalizes for a very large deviation from real distribution even with the addition of MI loss, thus finding a balance between accuracy and fairness (at $\alpha=0.5$).

		\section{Experimental Results and Discussions}
		
		{\bf Dataset}: We use MIMIC-III, a publicly available healthcare database containing de-identified patient admissions between 2001 and 2012 (33798 unique patient stays). We obtained permission to access MIMIC-III for research purposes after completion of an online course (certification number 45456719). More details regarding the cohort construction are given in Appendix.
		
		{\bf Benchmarks}: The methods we benchmark against are based on SOTA synthetic health data generation schemes and fair data generation. We compare with HealthGAN, as it is the best among the SOTA generative models in healthcare data. Next, we validate the performance against FairGAN on how the fairness measures have been improved in our settings. We could not find any references regarding fair data generation in healthcare, addressing the concerns raised in \cite{bhanot2021problem}. We do not include DECAF \cite{van2021decaf} as it is based on the underlying causality and finding causal inference knowledge on MIMIC-III is still a work-in-progress \cite{guzmanopen}, and we leave it for future study.
		
		For the ablation study, we use two variants namely Bt-GAN$^-$, which only satisfies $\mathcal U(\mathcal S, \mathcal Y)$, and Bt-GAN which accounts for both $\mathcal U(\mathcal S, \mathcal Y)$ and LDSS-based Dp-DGP. 
		
		{\bf Evaluation Metric}: We evaluate our proposed model using the following criteria:
		
		\begin{itemize}
			\item\textbf{Data Utility} - We use AUROC, AUPRC, F1 score, and accuracy for evaluating data utility. We train classifiers (Linear Regression (LR) and Random Forest (RF) on synthetic data for downstream tasks and validate the performance. Moreover, we perform sample-level metrics analysis proposed in \shortcite{alaa2022faithful} together with Jenson Shanon Divergence (JSD) \shortcite{sinn2018non} and discriminative score. 
			\item  \textbf{$\mathcal{U}(S, Y)$ Fairness} - This metric is used to measure the correlation bias and we use the parity gap and AUROC gap. Moreover, we perform data and model leakage \cite{wang2019balanced} to compare the bias amplification in our settings (Details in Appendix). 
			\item \textbf{Representation Fairness} - It is compared and evaluated by $LDSS$ calculated in every possible sub-groups. We use pie charts to plot the differences in representations captured by different models. 
		\end{itemize}
		
		{\bf Experimental Setup}: The details of different neural networks are given in Appendix. 
		\subsection{Data Utility Analysis}
		We focus on four types of binary prediction tasks for analyzing the data utility: (i) In-ICU mortality, (ii) In-hospital mortality, (iii) Length of Stay (LOS) greater than 3 days, and (iv) Length of stay (LOS) greater than 7 days. We use LR and RF models for predictions. Our aim here is to compare the quality and utility of the synthetic data generated by the proposed model, not the classifier accuracy.
		\begin{figure}[h]
			\centering
			\begin{subfigure}{0.45\textwidth}
				\includegraphics[width=\linewidth, height=4.5cm]{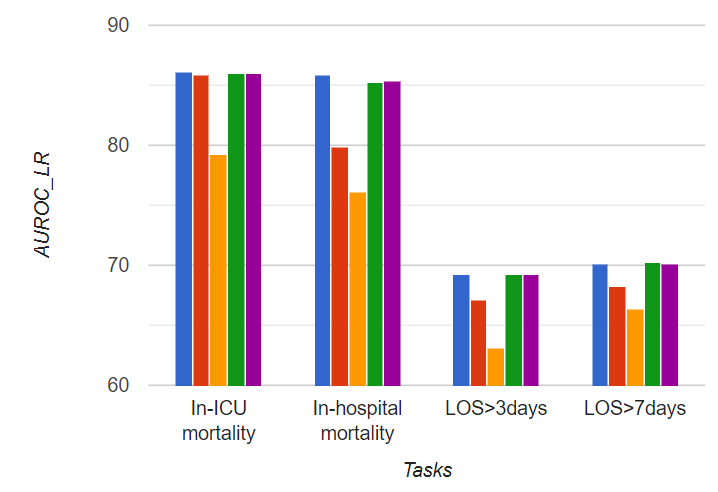} 
				\caption{AUROC\textunderscore LR}
				\label{}
			\end{subfigure}
			\begin{subfigure}{0.45\textwidth}
				\includegraphics[width=\linewidth, height=4.5cm]{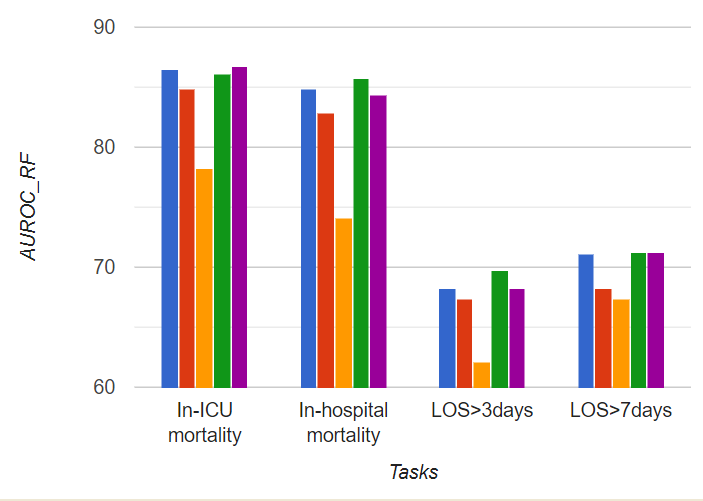}
				\caption{AUROC\textunderscore RF}
				\label{}
			\end{subfigure}
			\begin{subfigure}{0.45\textwidth}
				\includegraphics[width=\linewidth, height=6cm]{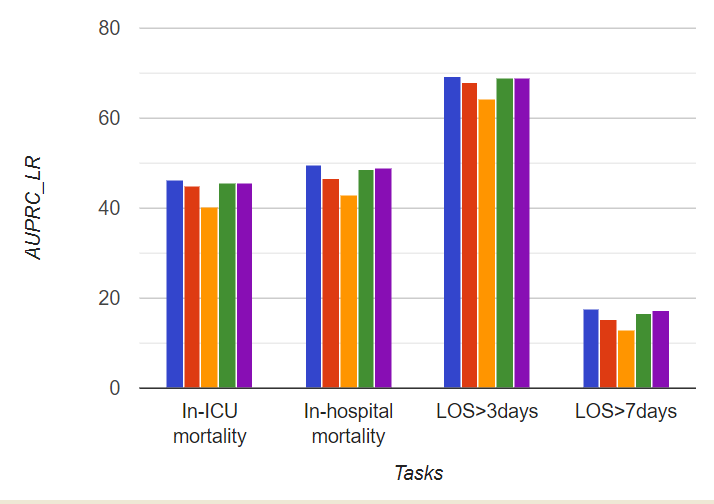}
				\caption{AUPRC\textunderscore LR}
				\label{}
			\end{subfigure}
			\begin{subfigure}{0.5\textwidth}
				\includegraphics[width=\linewidth, height=6cm]{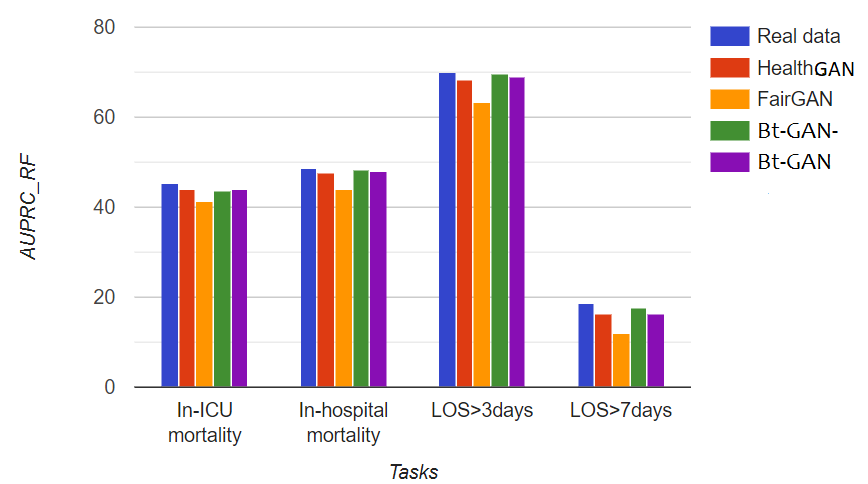}
				\caption{AUPRC\textunderscore RF}
				\label{}
			\end{subfigure}
			\caption{Data utility analysis}
			\label{fig:datautility}
		\end{figure}

		\begin{table}[hbt!]
			\centering
			{\small
				\begin{tabular}{ccccccccc}\toprule
					& \multicolumn{2}{c}{\textbf{Real Data}} & \multicolumn{2}{c}{\textbf{HealthGAN}} & \multicolumn{2}{c}{\textbf{FairGAN}} & \multicolumn{2}{c}{\textbf{Bt-GAN}}\\
					
					\cmidrule(lr){2-3} \cmidrule(lr){4-5} \cmidrule(lr){6-7} \cmidrule(lr){8-9}
					&accuracy. &F1 &accuracy. &F1 &accuracy. &F1 &accuracy. &F1 \\ 
					\midrule
					In-ICU mortality &92.1 &37.6 &91.3 &34.2 &89.5 &32.4 &\textbf {91.5} &\textbf{34.7} \\ 
					~  & 91.8 & 12.1 & \textbf{91.1} & \textbf{12.0} & 88.3 & 11.6 & 90.9 & 11.7 \\ 
					\midrule
					$LOS>3days$ & 71.2 & 59.9 & \textbf{69.4} & \textbf{58.3} & 67.1 & 56.3 & 68.1 & 57.3 \\ 
					~ & 72.6 & 59 & 67.2 & \textbf{59.1} & 66.4 & 57.9 & \textbf{68.9} & 57.6 \\
					\midrule
					In-hospital mortality & 90.1 & 39.6 & 89.1 & 37 & 85.4 & 32.8 & \textbf{89.6} & \textbf{39.9} \\ 
					~ & 89.3 & 17.9 & 88.3 & 15.8 & 86.3 & 14.3 & \textbf{90} & \textbf{18.1} \\ 
					\midrule
					$LOS>7days$ & 89.9 & 7.0 & 87.9 & \textbf{8.5} & 86.1 & 4.3 & \textbf{88.4} & 6.8 \\
					~ & 87.6 & 1.4 & \textbf{88.4} & 2.1 & 85.9 & 0.8 & 87.3 & \textbf{2.4} \\ \bottomrule
				\end{tabular}
			}
			\caption{Accuracy and F1 on various prediction tasks with real data as reference point. For each task, the first row denotes the predictions by LR, and the second row is the predictions by RF (higher is better for all the values).}
			\label{tab:acc-table}
		\end{table}
		
		

		\begin{figure}
			\centering
			\begin{subfigure}{0.44\textwidth}
				\includegraphics[width=\linewidth]{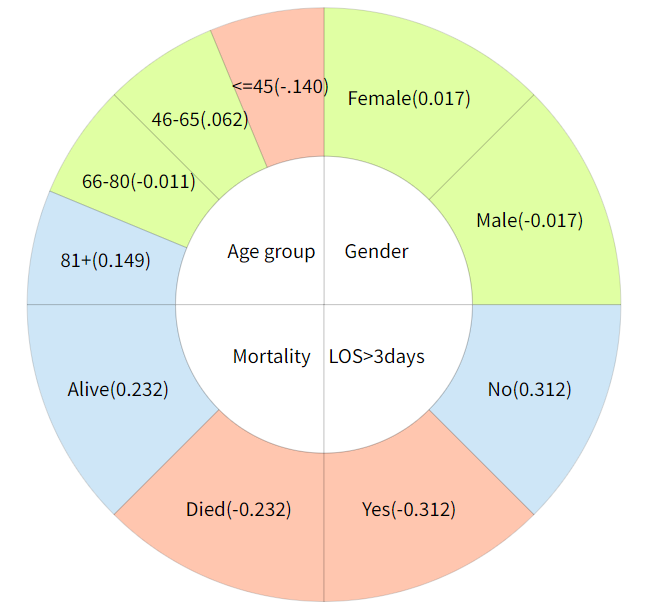} 
				\label{fig: real data}
			\end{subfigure}
			\begin{subfigure}{0.44\textwidth}
				\includegraphics[width=\linewidth]{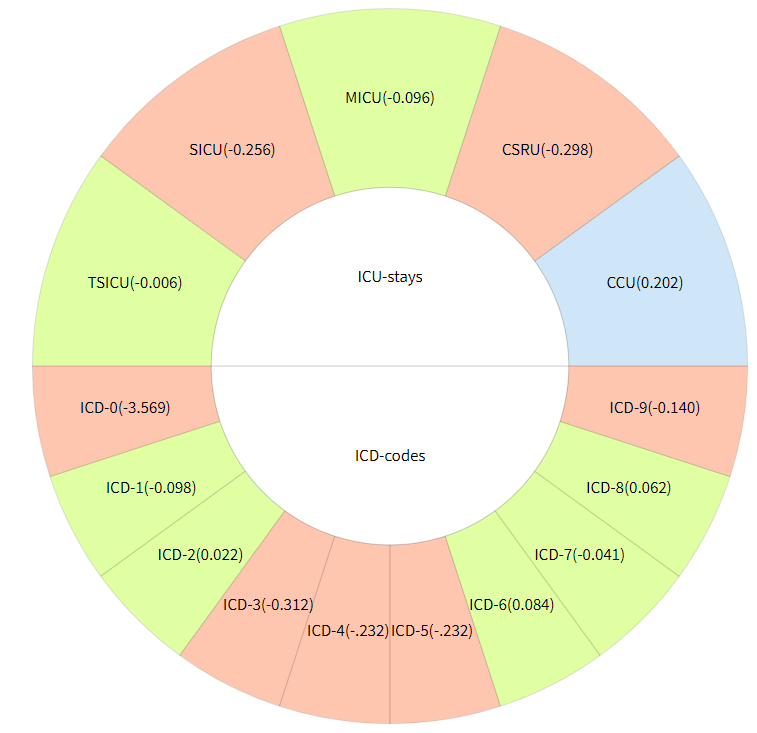}
				\small
				\caption{HealthGAN}
				\label{fig:health}
			\end{subfigure}
			\begin{subfigure}{0.44\textwidth}
				\includegraphics[width=\linewidth]{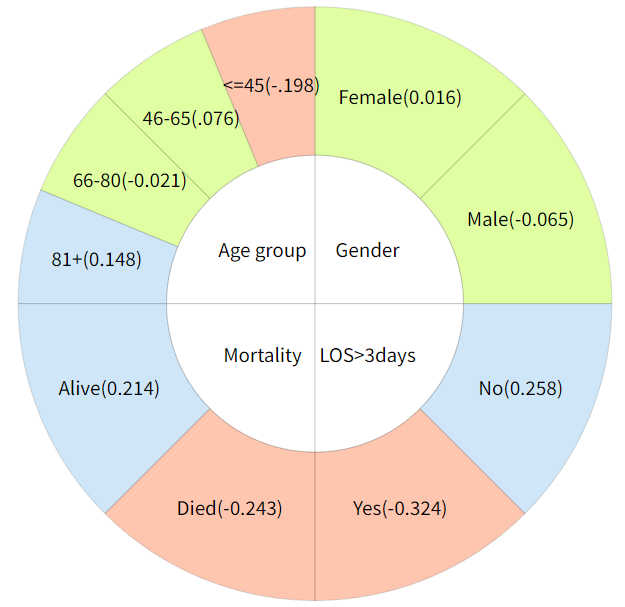}
				\small
				\label{fig:HealthGAN}
			\end{subfigure}
			\begin{subfigure}{0.44\textwidth}
				\includegraphics[width=\linewidth]{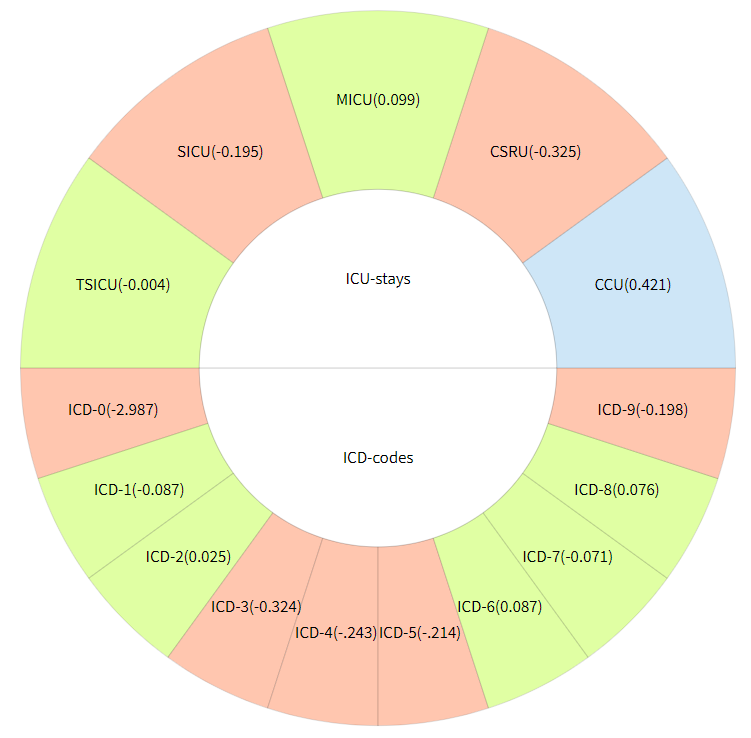}
				\small
				\caption{Bt-GAN $^-$}
				\label{fig:DeMISe}
			\end{subfigure}
			
			\begin{subfigure}{0.44\textwidth}
				\includegraphics[width=\linewidth]{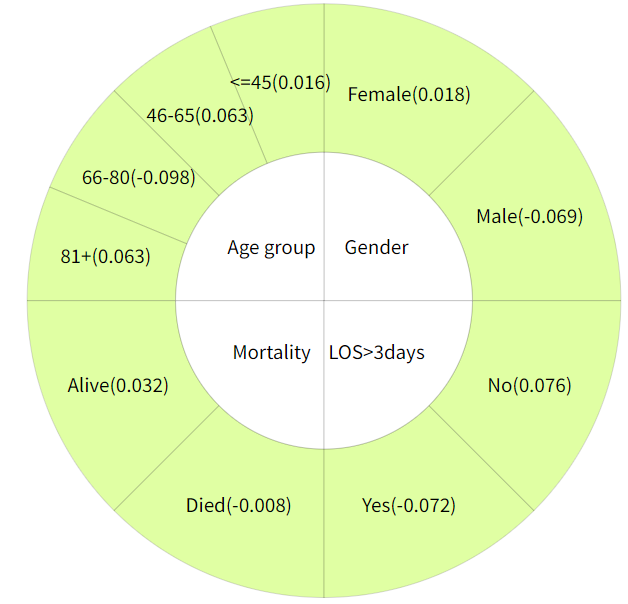}
				\small
				\label{fig:DeMISe}
			\end{subfigure}
			\begin{subfigure}{0.44\textwidth}
				\includegraphics[width=\linewidth]{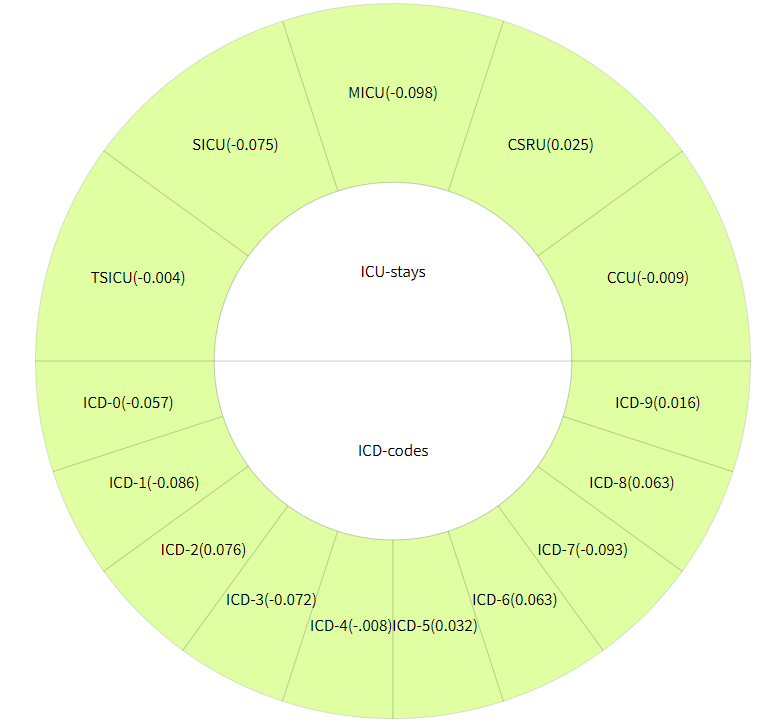}
				\small
				\caption{Bt-GAN}
				\label{fig:DeMISe}
			\end{subfigure}
			\caption{Representation of sub-groups (The LDSS scores are given in brackets) }
			\label{fig:representations}
		\end{figure}
		
		\begin{figure}[hbt!]
			\centering
			\includegraphics[width=0.8\linewidth, height=11cm]{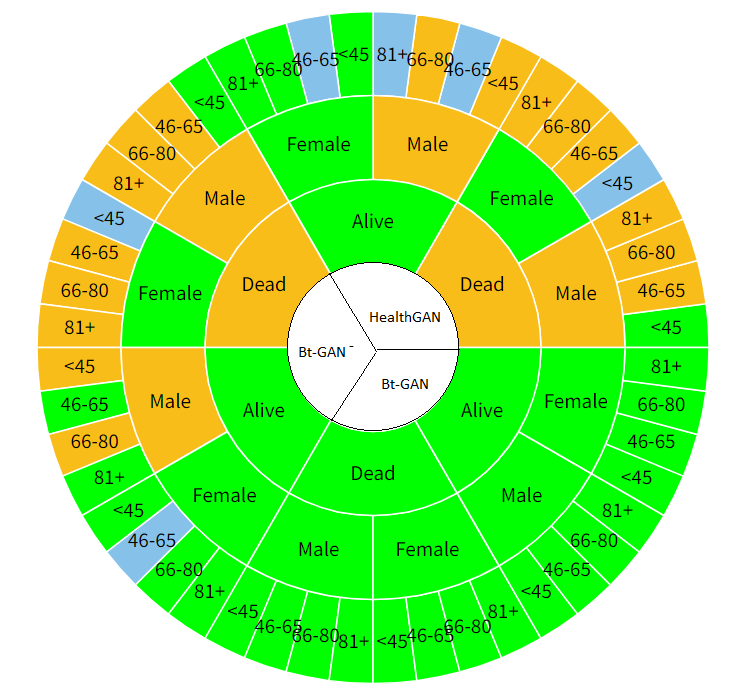}
			\caption{Comparison of log disparity values on the combination of attributes such as mortality, age, and gender. The underrepresented (orange), over-represented (blue), and adequately represented (green) demographic combinations of the proposed models are compared with HealthGAN. The chart areas are respectively divided as models, mortality, gender, and age. }
			\label{fig:comparisonlog}
		\end{figure}

		\begin{figure}[h]
			\centering
			\begin{subfigure}{0.45\textwidth}
				\includegraphics[width=\linewidth, height=6cm]{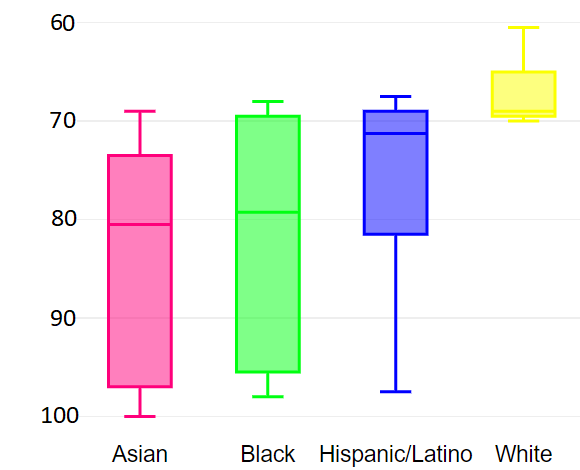} 
				\caption{Real data}
				\label{fig:real data}
			\end{subfigure}
			\begin{subfigure}{0.45\textwidth}
				\includegraphics[width=\linewidth, height=6cm]{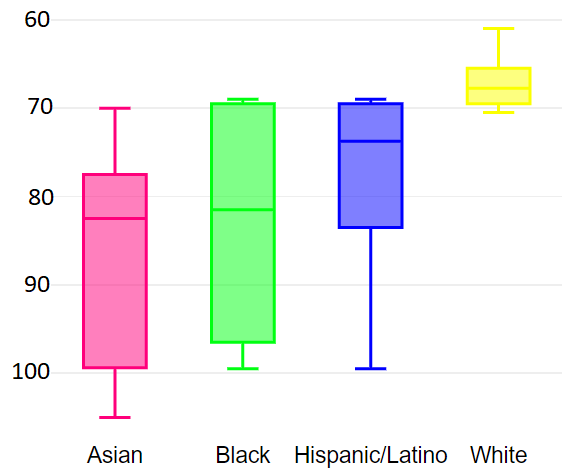}
				\caption{HealthGAN}
				\label{fig:HealthGAN}
			\end{subfigure}
			\begin{subfigure}{0.45\textwidth}
				\includegraphics[width=\linewidth, height=6cm]{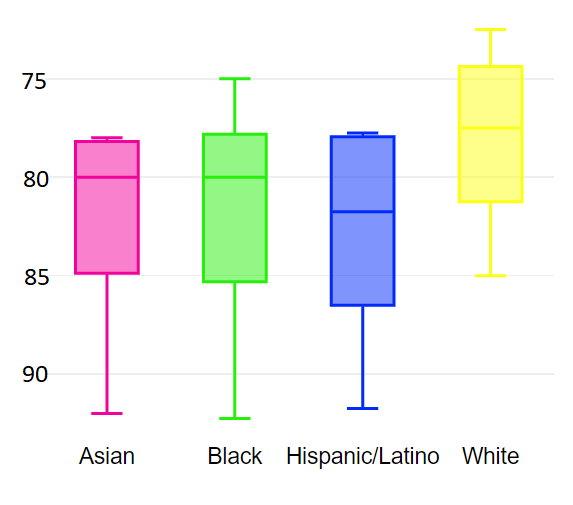}
				\caption{FairGAN}
				\label{fig:DeMISe}
			\end{subfigure}
			\begin{subfigure}{0.45\textwidth}
				\includegraphics[width=\linewidth, height=6cm]{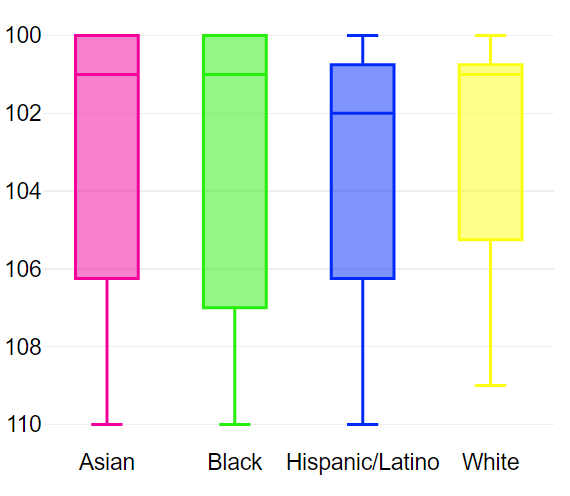}
				\caption{Bt-GAN}
				\label{fig:DeMISe}
			\end{subfigure}
			\caption{Importance of \textit{ethnicity} among sub-groups by LR using SHAP (feature ranks in the vertical axis) }
			\label{fig:explainability}
		\end{figure}

		Table \ref{tab:acc-table} reports the accuracy and F1 score of the four downstream tasks. The data generated by HealthGAN and Bt-GAN(ours) maintain the accuracy and F1 score for all the prediction tasks as that of real data, whereas FairGAN fails to keep those. The performance of our model is commendable as it keeps almost the same accuracy and F1 score as that of HealthGAN with added fairness. Additionally, we compare utility in terms of AUROC and AUPRC using LR and RF. Figure \ref{fig:datautility} shows 4 plots: (a) AUROC\textunderscore LR, (b) AUROC \textunderscore RF, (c) AUPRC\textunderscore LR, and (d) AUPRC\textunderscore RF. The difference between HealthGAN and Bt-GAN is negligible in all plots, whereas FairGAN achieves the worst performance. Finally, we provide sample-level metrics analysis \cite{alaa2022faithful} to verify the quality, fidelity, diversity, and generalization in Table \ref{tab:quantative}. Our model is superior in the discriminative score, $\beta$ recall, JSD, authenticity, and context FID compared to SOTA.
		\begin{table*}[!h]
			\centering
			{\small
				\begin{tabular}{ccccccc} \toprule
					\textbf{Model} &\textbf{Discriminative} &\textbf{JSD} &$\alpha$ &$\beta$  &\textbf{Authenticity} &\textbf{Context} \\ 
					&\textbf{score} & &\textbf{precision} &\textbf{recall} & &\textbf{FID}\\
					& ($\downarrow$) & ($\downarrow$) & ($\uparrow$)& ($\uparrow$)& ($\uparrow$)& ($\downarrow$)\\ \midrule
					
					HealthGAN & 0.31$\pm$.002 &0.032$\pm$.01 &\textbf{0.82$\pm$.002} &0.52$\pm$.003 &0.91$\pm$.001 &0.89$\pm$.021\\ 
					\midrule
					FairGAN & 0.46$\pm$.24 &0.074$\pm$.05 &0.56$\pm$0.53 &0.21$\pm$0.12 &0.62$\pm$.001 &3.12$\pm$.32 \\
					\midrule
					Bt-GAN$^-$  & 0.35 $\pm$0.01  & \textbf{0.03$\pm$.21} & 0.70$\pm$.15 & 0.54$\pm$.04 & 0.88$\pm$.13 & 0.89$\pm$.13 \\ 
					
					Bt-GAN  &\textbf{0.29 $\pm$0.01} &0.031 $\pm$.001 &0.810$\pm$.001 &\textbf{0.61 $\pm$0.032} &\textbf{0.92$\pm$.010} &\textbf{0.81$\pm$.12}\\
					\bottomrule
				\end{tabular}
			}
			\caption{Quantitative analysis;$\uparrow$ indicates higher the better, $\downarrow$ indicates lower the better, best results are bolded).}
			\label{tab:quantative}
		\end{table*}
		
		\subsection{Fairness Analysis}
		MIMIC-III is composed of a set of sensitive features. As per Equal Credit Opportunity Act [ECOA], gender, age, ethnicity, insurance type, and marital status are considered sensitive information. To compare the fairness of the proposed model with SOTA, we choose \textit{ethnicity} on which we enforce $\mathcal U(\mathcal S, \mathcal Y)$ fairness. Though there are no restrictions in choosing the protected features in this study, enforcing equality on age influences the patient’s health which may affect mortality or length of stays in the ICU and cause unnecessary medical interventions. So, we recommend the choice of protected features based on the context of applications and suggestions from healthcare experts.
		
		To compare $\mathcal U(\mathcal S, \mathcal Y)$ fairness in ethnicity sub-groups, we choose two prediction tasks by LR: (i) In-hospital mortality, and (ii) $LOS>7days$ on which we analyze the parity and AUROC gap between black and white patients across various synthetic data.
		
		Table \ref{tab:comparisonoffairness} compares the parity as well as AUROC gaps in different GANs with real data. The AUROC gap and the parity gap show that predictions on HealthGAN-generated data are biased and amplified compared to real data. In all the cases, a positive value represents a bias towards white patients and a negative value represents a bias towards black patients. The data is fair when the value is close to zero. FairGAN minimizes these biases by adversarial debiasing. The gaps in Bt-GAN are almost close to zero for all the prediction tasks considered. With this being said, the differences in parity, as well as AUROC gaps between HealthGAN-generated data and Bt-GAN, are significant for these predictions.  
		
		\begin{table*}[]
			\centering
			{\small
				\begin{tabular}{cccccc}\toprule
					\textbf{Metrics} & \textbf{Prediction}  & \textbf{Real Data}     & \textbf{HealthGAN} & \textbf{FairGAN} & \textbf{Bt-GAN}       \\ \midrule                             
					& In-hospital mortality  & 0.043 $\pm$ 0.001 & 0.082 $\pm$ 0.002 & 0.021 $\pm$ 0.001  & \textbf{0.001 $\pm$ 0.001}  \\
					AUROC  & In-ICU mortality  & 0.03 $\pm$ 0.007 & 0.15 $\pm$ 0.035 & 0.023 $\pm$ 0.064 & \textbf{0.012 $\pm$ 0.021}\\
					gap  & LOS\textgreater{}3days & -0.003 $\pm$ 0.002  & -0.104 $\pm$ 0.001 & -0.003 $\pm$ 0.001 & \textbf{0.000 $\pm$ 0.001}  \\
					& LOS\textgreater{}7days & -0.005 $\pm$ 0.002 & -0.076 $\pm$ 0.002 & -0.061 $\pm$ .001 & \textbf{-0.013 $\pm$ 0.001} \\
					\midrule
					& In-hospital mortality& -0.046 $\pm$ 0.018  & -0.154 $\pm$ 0.010 & -0.004 $\pm$ 0.014  & \textbf{0.000 $\pm$ 0.001}  \\
					Parity  & In-ICU mortality & -0.031 $\pm$ 0.013 & -0.331 $\pm$ 0.011 & -0.005 $\pm$ 0.013 & \textbf{0.000 $\pm$ 0.000}  \\
					gap & LOS\textgreater{}3days & 0.022 $\pm$ 0.012 & 0.224 $\pm$ 0.012 & 0.022 $\pm$ 0.002 & \textbf{0.000 $\pm$ 0.001}  \\
					& LOS\textgreater{}7days & -0.004 $\pm$ 0.002 & -0.004 $\pm$ 0.002 & \textbf{-0.002 $\pm$ 0.001} & -0.003 $\pm$ 0.001 \\ 
					\bottomrule
				\end{tabular}
			}
			\caption{Comparison of fairness gaps between white and black patients with real data as reference (best results are in bold).}
			\label{tab:comparisonoffairness}
		\end{table*}
		
		To further evaluate the data leakage and model leakage \cite {wang2019balanced} of our proposed model, we train an attacker (which is an \textit{ethnicity} classifier) on the ground truth labels (of various generated data) and the corresponding downstream predictions (by LR)  of the In-ICU mortality task. Note that the leakage of LR on the real dataset is $0.60\pm001$ with an F1 score of 92.3. This shows that the underlying generative model amplifies these leakages as shown in Table \ref{tab:evaluationofdata}. Our proposed Bt-GAN models achieve superior performance in controlling these leakages as the bias amplification factor $\Delta$, which is the difference between the model and data leakage is less than zero (details in Appendix). Note that adversarial de-biasing by FairGAN does not account for both leakages, but is mitigated in Bt-GAN through MI-debiasing in the Bt-DGP. 
		
		\textit{Remark}. The adversarial de-biasing-based generation process generates fair data by fooling the discriminator. Our Bt-GAN method uses a module for estimating the mutual information which does not compete with the generator. That means we generate fair health data without fooling the estimator, but by minimizing the information it estimates. The advantage of our method is that we can train the estimator until convergence in every epoch which is not possible in adversarial de-biasing methods.
		
		\begin{table}[h!]
			\centering
			{\small
				\begin{tabular}{ccc}\toprule
					\textbf{Data} &\textbf{Data leakage} & \textbf{Model leakage} \\ 
					\midrule
					HealthGAN &63.17$\pm$0.45 &65.42$\pm$0.58\\
					FairGAN &60.81$\pm$0.32 &62.97$\pm$0.31\\
					Bt-GAN$^-$ &50.45$\pm$0.10 &49.16$\pm$0.14\\
					Bt-GAN &\bf{49.90$\pm$0.63} &\bf{48.31$\pm$0.71}\\
					\bottomrule
				\end{tabular}
			}
			\caption{Evaluation of data and model leakage. }
			\label{tab:evaluationofdata}
		\end{table}
		
		\subsection{Representation Fairness Analysis}
		\label{sub:reprefair}
		It is essential for health sector research to incorporate demographic information as part of clinical research. One such example could be to test the effect of vaccines on different age groups, gender, or different physical conditions. So, the real data should include appropriate proportions of all the sub-groups to use for clinical research effectively. Thus, similar distributions of sub-groups should be captured when it is being generated.
		
		Our analysis of representation fairness can be performed in two ways: (i) on sub-groups identified by gender, age, in-hospital mortality, and $LOS>3days$, and (ii) on sub-groups defined by ICD-codes and ICU stay in the emergency care unit. Note that, these sub-groups are selected not based on any clinical research but to analyze how these proportions are being propagated into synthetic data. 
		
		Similar to \cite{bhanot2021problem}, we set the value of $\delta$ (Definition 9) to $\pm \log (0.9)$ (90 percent rule) as a threshold for over/underrepresented groups. Based on this, we divide the range into 4 levels, such as missing $[-\propto$ to $\log (.8)]$ (yellow), underrepresented $(\log (.8)$ to $\log (.9)]$ (orange), adequately represented $(\log(.9)$ to $-\log (.9)]$ (green) and over-represented $(-\log(.9)$ to $-\log (.8)]$ (blue).
		
		\textbf {Analysis of Sub-groups Identified by Gender, Age, In-hospital Mortality, and $LOS>3days$. } We analyze the individual representation issues and the issues caused by the intersection of these sub-groups in detail. Figure \ref{fig:representations} (top row) compares representation issues calculated by $LDSS$ for each of the sub-groups.
		More representation biases (under/over) are caused by HealthGAN and Bt-GAN$^-$-generated data.  Specifically, more over-representation can be seen towards the attributes defined by  $age='81+'$, $mortality='Alive'$, and $LOS>3days='No'$. The attributes $age='<=45'$, $mortality='Died'$, and $LOS>3days='Yes'$ are underrepresented. Suppose, if the majority of the under-represented combinations belong to one particular class and the over-represented in the other, the predictions by a downstream classifier will be severely biased, even though the overall prediction accuracy is acceptable.

		\textbf {Analysis of Sub-groups Identified by ICD Codes and ICU Stay in the Emergency Care.} We analyze the representation biases caused by ICD codes and various ICU stays (CCU, CSRU, MICU, SICU, and TSICU) in Figure \ref{fig:representations} (bottom row). This analysis is very important in healthcare for setting up emergency care based on appropriate disease characteristics (defined by ICD codes). Among the ICU stays, CSRU and SICU are underrepresented in both the data generated by HealthGAN and Bt-GAN$^-$, whereas CCU is over-represented. The ICD codes (ICD-0,3,4,5, and 9) are underrepresented in HealthGAN and Bt-GAN$^-$ (more results are in Appendix). Note that, Bt-GAN captures all these sub-groups in exact proportions in contrast to other schemes. Furthermore, we analyze the representation issues caused by the cross-section of sub-groups in Figure \ref{fig:comparisonlog}. Though the representations of gender are correctly captured by all the models, their intersections with other demographics cause certain biases.
		
		\subsection{Attribute-based Local Explainability Analysis using SHAP}
		Motivated by \shortcite{cesaro2019measuring}, we analyze the group feature importance for samples in In-hospital mortality prediction by LR and measure how attributed it is across various sub-groups defined by the \textit{ethnicity} group. We then rank the importance among other features within  SHAP exploratory analysis as shown in Figure \ref{fig:explainability}. We observe that the attribute \textit{ethnicity} has a great impact (low ranks) on the sub-groups \textit{White} in real data, FairGAN, and HealthGAN resulting in biased explanations \shortcite{jain2020biased}. In Bt-GAN, the effect is more balanced and even reduced, which means the feature \textit{ethnicity} is less important to the model for all the sub-groups. This local analysis is helpful in finding the disparities between the sub-groups in situations where global explanations cannot reveal the inequalities between the sub-groups. 
		
		\section{Experiment on Fairness Benchmark Datasets}
		In this section, we evaluate the effectiveness of our model in producing fair synthetic data on fairness benchmark datasets such as Adult Income\footnote{https://archive.ics.uci.edu/ml/datasets/adult} and COMPAS\footnote{https://www.propublica.org/article/machine-bias-risk-assessments-in-criminal-sentencing}. Note that the missing labels in both of these datasets are negligible. Therefore, semi-supervised learning has no impact on the generation quality, and thus the terms associated with the classifier in Eq. (4) have no relevance. As a result, Eq. (4) can be replaced by Eq. (3) for any datasets with no (or negligible) missing labels. 
  
		\subsection{Datasets} \paragraph{UCI Adult Dataset} This dataset is based on US census data (1994) and contains 48,842 rows with attributes such as age, sex, occupation, and education level, and the target variable indicates whether an  individual has an income that exceeds $50K$ per year or not. In our experiments, we consider the protected attribute to be sex (S = “Sex”, Y = “Income”).
		
		\paragraph{ProPublica Dataset from COMPAS Risk Assessment System} This dataset contains information about defendants from Broward County and contains attributes about defendants such as their ethnicity, language, marital status, sex, etc., and for each individual a score showing the likelihood of recidivism (reoffending). In this experiment, we used a modified version of the dataset. First, attributes such as FirstName, LastName, MiddleName, CASE ID, and DateOfBirth are removed. Studies have shown that this dataset is biased against African Americans. Therefore, ethnicity is chosen to be the protected attribute for this study. Only African American and Caucasian individuals are kept and the rest are dropped. The target variable in this dataset is a risk decile score provided by the COMPAS system, showing the likelihood of that individual to re-offend, which ranges from 1 to 10. The final modified dataset contains 16,267 records with 16 features. To make the target variable binary, a cut-off value of 5 is considered and individuals with a decile score of less than 5 are considered “Low Chance”, while the rest are considered “High Chance”. (S = “Ethnicity”, Y = “Recidivism Chance”).

		\paragraph{Competing Methods}
		The methods we compare against are based on fair tabular data generation using GANs. We compare against FairGAN and DECAF as these are state-of-the-art methods in the tabular domain. 
					
					
					

		\begin{table*}[]
			\centering
			{\small
				\begin{tabular}{cccccc}\toprule
					& Datasets & Precision & Recall &AUROC gap &Parity gap \\ \midrule
					& Adult & 0.902 $\pm$ 0.001 & 0.921 $\pm$ 0.002&0.198 ± 0.018&0.121 ± 0.024 \\
					Real Data & COMPAS & 0.903 $\pm$ 0.007 & 0.914 $\pm$ 0.007&0.239$\pm$ 0.002 & 0.258 $\pm$0.032\\ \hline
					& Adult &   0.781 $\pm$ 0.018 & 0.881 $\pm$ 0.050&0.011 ± 0.002 &0.001 ± 0.001 \\
					DECAF & COMPAS &  0.874 $\pm$ 0.010& 0.886 $\pm$ 0.001&0.021 ± 0.010 &0.003 ± 0.011\\ \hline
					& Adult & 0.661 $\pm$ 0.140 & 0.679 $\pm$ 0.001 &0.089$\pm$0.002& 0.097 ± 0.018\\
					FairGAN & COMPAS &  0.785 $\pm$ 0.010 & 0.832 $\pm$ 0.002&0.045$\pm$0.023&0.205 ± 0.055\\ \hline
					& Adult & \textbf{0.898} $\pm$ \textbf{0.001} & \textbf{0.900} $\pm$ \textbf{0.020}&\textbf{0.002$\pm$ 0.010}&\textbf{0.001$\pm$ 0.020}\\
					Bt-GAN (ours) & COMPAS & \textbf{0.896} $\pm$ \textbf{0.001} &\textbf{0.899} $\pm$ \textbf{0.003}&\textbf{0.001 $\pm$ 0.002}&\textbf{0.002$\pm$ 0.001}\\ \bottomrule
					
				\end{tabular}
			}
			\caption{Data utility and fairness analysis on fairness benchmark datasets with real data as the reference point (higher is better for all the values)}
			\label{tab:accf1}
		\end{table*}
		\subsection{Results}
		We list the utility and fairness measures in Table \ref{tab:accf1}. The precision and recall of our proposed Bt-GAN are better than FairGAN and DECAF.  The fairness evaluated in terms of the AUROC gap and parity gap of Bt-GAN shows that the bias in synthetic data has been reduced to a great extent compared to state-of-the-art models. We achieve a good balance between utility and fairness through a Bt-DGP. 
		
		\section{Conclusion and Future Works}
		
		We proposed Bt-GAN, a semi-supervised Generative Adversarial Network for synthetic healthcare data generation, which accounts for the spurious correlations in the data and promotes representation fairness in the target in an end-to-end framework. This is done by proposing a Bias-transforming generation process, incorporating an information-constrained fairness penalty to tackle correlation biases, a score-based weighted sampling to promote representation fairness, and semi-supervised learning to capture true data distribution under partially unlabeled data. Compared to the SOTA schemes, our method is found to be more reliable to better balance the tradeoff between accuracy and fairness with minimal bias amplification in a more generalized and principled way. 
		
		{\bf Future Directions:} This work is based on GANs and feature-specific constraints such as mutual information. An interesting future direction could be to extend this in the framework using diffusion models and causality. We leave further experiments to extend our proposed methods from these perspectives as future work. 
		
		{\bf Ethical and Societal Implications:} The notion of fairness is domain and context-dependent, which is more sensitive in healthcare. The proposed methods could be coupled with tasks, where the goal is the progression of trustworthy AI.
		

        \acks{This work was funded by the Knut and Alice Wallenberg Foundation, the ELLIIT Excellence Center at Linköping-Lund for Information Technology, and TAILOR - an EU project with the aim to provide the scientific foundations for Trustworthy AI in Europe. The computations were enabled by the Berzelius resource provided by the Knut and Alice Wallenberg Foundation at the National Supercomputer Centre.} 	

          \appendix

		\section{Quality and Fairness Definitions}
		
		\paragraph{Discriminative Score:} To calculate the score, we train a classifier(LR) to classify \textit{real} or \textit{fake} with the original data and synthetic data respectively labeled as real and fake and calculated the error.
		
		\paragraph{Jensen-Shannon Divergence (JSD):} Here we used JSD to evaluate the quality of the synthetic data as compared to the real data \shortcite{menendez1997jensen}.
		
		\paragraph{$\alpha$-precision, $\beta$-recall, and Authenticity:} It is used to evaluate the fidelity, diversity, and generalization of the synthetic data at the sample level as proposed in \shortcite{alaa2022faithful}.
		
		\paragraph{Context FID:} It is used to measure the difference in statistics between the real and synthetic data with respect to downstream tasks \shortcite{jeha2021psa}.
		
		\paragraph{Data Leakage:} It is used to evaluate how much information about the protected attributes can be leaked through task-specific labels. To measure this, we train an attacker $f$ and evaluate it on held-out data. The performance of the attacker, the fraction of instances in $\mathcal{D}$ that leak information about $S_{i}$ through $Y_{i}$, yields an estimate of leakage \cite{wang2019balanced}:
		\begin{equation}
			Leakage_{\mathcal{D}}=\frac{1}{|\mathcal{D}|} \sum_{\left(Y_{i}, S_{i}\right) \in \mathcal{D}} {1}\left[f\left(Y_{i}\right)==S_{i}\right]
		\end{equation}
		
		\paragraph{Model Leakage:} To measure the degree a model, $M$ produces predictions, $\hat{Y}_{i}=M\left(X_{i}\right)$, that leak information about the protected variable $S_{i}$. We define model leakage as the percentage of examples in $\mathcal{D}$ that leak information about $S_{i}$ through $\hat{Y}_{i}$. To measure prediction leakage, we train a different attacker on $\hat{Y}_{i}$ to extract information about $S_{i}$ \cite{wang2019balanced} :
		\begin{equation}
			Leakage_{M}=\frac{1}{|\mathcal{D}|} \sum_{\left(\hat{Y}_{i}, S_{i}\right) \in \mathcal D} {1}\left[f\left(\hat{Y}_{i}\right)==S_{i}\right)]
		\end{equation}
		
		We measure fairness using two measures; the AUROC gap and the parity gap. The AUROC gap is defined as the difference in AUROC between the selected sub-groups for a particular healthcare task. The parity gap is the difference in statistical parity between two sub-groups for a specified task. In Table \ref{tab:fairnessdefinition}, we give definitions of the AUROC gap and Parity gap for sub\textunderscore{groups} $SG_{i}$.
		\begin{table}[H]
			
			\centering
			\begin{tabular}{l|l}\toprule
				
				AUROC gap     &$\operatorname{AUROC}\left(\mathrm{SG}_{1}\right)-\mathrm{AUROC}\left(\mathrm{SG}_{2}\right)$     \\
				\midrule
				
				Parity Gap     &$\frac{T P_{1}+F P_{1}}{N_{1}}-\frac{T P_{2}+F P_{2}}{N_{2}}$\\   
				\bottomrule
			\end{tabular}
			\caption{Fairness definitions}
			\label{tab:fairnessdefinition}
		\end{table}

		\section{Extensions to Other Fairness Notions}
		
		Analogous to the connection between statistical parity and zero mutual information, the fairness notions such as equalized opportunity and equalized odds can be designed as a conditional mutual information optimization problem as detailed in \cite{ghassami2018fairness}.
		
		\section{Architecture Details}
		
		\paragraph{Implementation Details:} The training epochs use a mini-batch size of 1024. The learning rate is set to .0001 with Adam optimizer. We carried out the experiments using PyTorch in Intel Core i-9, 11th generation, with 128 GB RAM and GPU-2* NVIDIA RTX 2080TI (11 GB). Also, the details architecture used in this study can be found on Table \ref{tab:neuralnetwork}.

		\begin{figure}[h]
			\centering
			\begin{subfigure}{0.4\textwidth}
				\includegraphics[width=\linewidth, height=5cm]{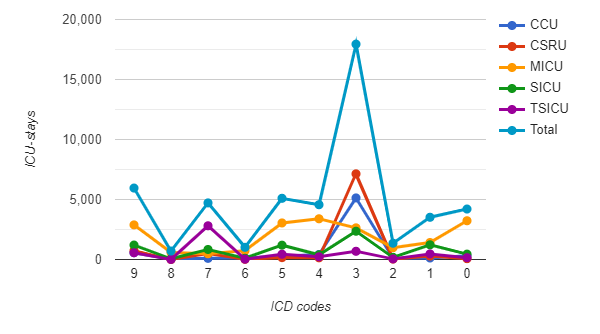} 
				\caption{Real data}
				\label{fig:real data}
			\end{subfigure}
			\begin{subfigure}{0.4\textwidth}
				\includegraphics[width=\linewidth, height=5cm]{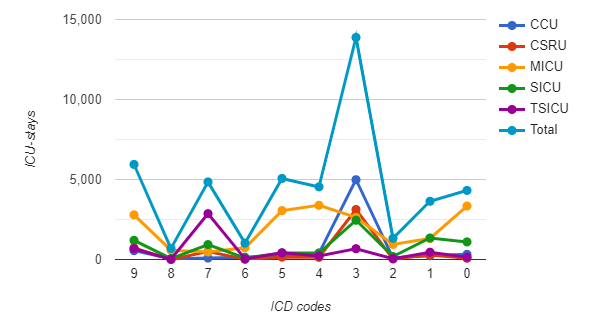}
				\caption{HealthGAN}
				\label{fig:HealthGAN}
			\end{subfigure}
			\begin{subfigure}{0.4\textwidth}
				\includegraphics[width=\linewidth, height=5cm]{line-graph3-1.png}
				\caption{Bt-GAN-}
				\label{fig:Bt-GAN}
			\end{subfigure}
			\begin{subfigure}{0.4\textwidth}
				\includegraphics[width=\linewidth, height=5cm]{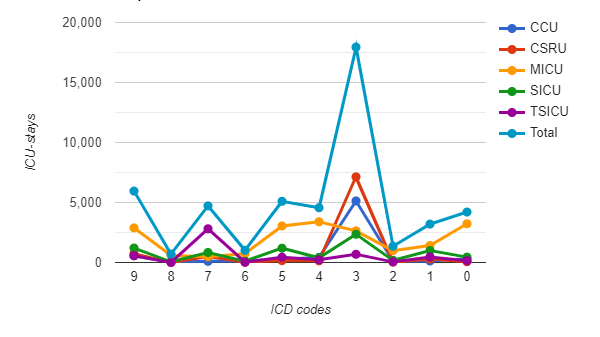}
				\caption{Bt-GAN}
				\label{fig:Heatmap}
			\end{subfigure}
			\caption{Distribution of ICD-codes in ICU stays}
			\label{fig:distributionICD}
		\end{figure}
		
		\begin{figure}[!h]
			\includegraphics[width=1\linewidth,height=5cm]{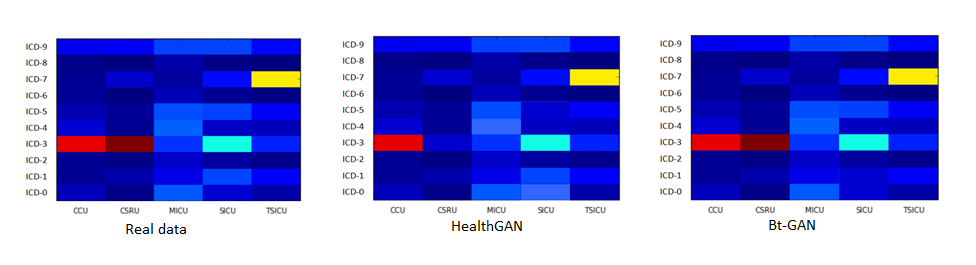}
			\caption{Heatmap representation of ICD codes and ICU stays in real data, HealthGAN generated data and Bt-GAN generated data }
			\label{fig:heatmap}
		\end{figure}
		
    \paragraph{Dataset Preparation:}
		We extracted records from tables, PATIENTS, ADMISSIONS, ICU STAYS, CHARTEVENTS, LABEVENTS, and OUTPUTEVENTS. Records are then validated using HADM\textunderscore{ID} and ICUSTAY\textunderscore{ID}, resulting in a total of 33,798 patients with 42276 ICU stays. Among them, we split 28728 patients, 35948 ICU stays for training, and the remaining for testing. We excluded patients with missing HADM\textunderscore{ID} and ICUSTAY\textunderscore{ID}. 
		
		\section{Additional Results}
		Additional results on the representation analysis are given in Figure \ref{fig:distributionICD} and Figure \ref{fig:heatmap}.
		
		\begin{figure}[!h]
			\centering
			\includegraphics[width=0.6\linewidth, height=4cm]{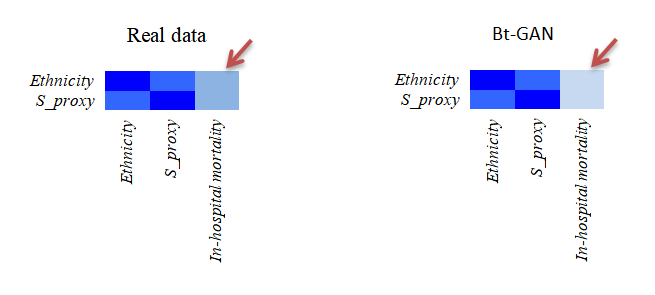}
			\caption{The MI de-biasing accounts for proxy attribute}
			\label{fig:MIdebiasing}
		\end{figure}

		\section {Cohort Summary}
		Table \ref{tab:cohortsummary} details the cohort summary defined by various demographic information.
		\begin{table}[!h]
			\centering
			\begin{tabular}{ccc}

				{
					Attributes}                                                    & Groups          & Percentage \\
				\hline
				
				&  Black      &  8\%      \\
				& White     &  71\%       \\
			Ethnicity	&  Hispanic&3\%       \\
				& Other     & 15\%       \\
				& Asian      & 2\%       \\
				\hline
				
				&
				Female    & 43\%      \\
				Gender    &  Male       & 57\%      \\
				\hline
				
				& Self-pay   & 1\%        \\
				
				& Government & 3\%        \\
			Insurance	&  Medicaid  &  8\%        \\
				
				& Private    & 34\%      \\
				
				 & Medicare   & 53\%      \\
				\hline
				& TSICU      & 13\%       \\
				& CCU                               & 15\%                              \\
			Emergency	& SICU                              & 16\%                              \\
				& CSRU                              & 20\%                              \\
				 & MICU                              & 35\%                              \\
				\hline
				Total                                                                      &                                   & 33798(100\%)\\
				\hline
			\end{tabular}
			\caption{Cohort summary}
			\label{tab:cohortsummary}
		\end{table}
		\begin{table*}[hbt!]
			\centering
			\begin{tabular}{cccc}
				\hline
				$\mathrm{C}_{\varphi}$ &$G_{\theta}$ &$\mathrm{D}_{\phi}$ , $D_{\zeta}$ &$T_{\eta}$ \\ 
				
				\hline
				z & No. of features &$(W, S)$ & $-T_{n}(\mathrm{~W})$\\
				(normal distribution) & - Dense layer of 64 &- Dense layer of 64 & X - Dense layer of  8\\
				- Dense layer of 2 & - leaky ReLU &- leaky ReLU  &- ReLU\\
				× no.of features -ReLU & - Dense layer of 128 &- Dense layer of 64 &- Dense layer of  8\\
				BN - Dense layer of 1.5 & - leaky ReLU &- leaky ReLU &- ReLU\\
				× no.of features- ReLU &- Dense layer of 256 &- Dense layer of 64 &- Dense layer of 2 \\
				- Dense layer of 1 & - leaky ReLU &- leaky ReLU &- ReLU -  Y \\
				× no.of features- ReLU &- Dense layer of 1 & & \\
				- Gumble$_{0.2}$ softmax(discrete) &- leaky ReLU & &\\  
				\hline
			\end{tabular}
			
			\caption{Neural Networks for Generator, $G_\theta$, Discriminator, $D_\phi$, $D_\zeta$, Classifier, $C_\varphi$, and MINE, $T_\eta$}
			\label{tab:neuralnetwork}
		\end{table*}
		\section{Impact of Proxy Attributes}
		We experimented further to study the impact of proxy attributes to check whether these can introduce any biases. We added an extra feature, \textit{$S\textunderscore{proxy}$} that is strongly correlated with \textit{ethnicity}, particularly for white and black sub-groups. For black patients, $S\textunderscore{proxy}=1$ for 95 percent of all cases and $S\textunderscore{proxy}=0$ for the remaining 5 percent. For white patients, the above values are swapped. A correlation plot shows that there is a strong correlation between \textit{ethnicity} and $S\textunderscore{proxy}$ as well as the combination of these to the mortality in the real health data. In our synthetic data, the correlation between 'ethnicity' and $S\textunderscore{proxy}$ remains the same but the correlation of both of these to the mortality is reduced to a great extent. So, it is evident that the MI de-biasing accounts for proxy attributes as well (Figure \ref{fig:MIdebiasing}).

		\bibliography{main}
		\bibliographystyle{theapa}

	\end{document}